\documentclass[letterpaper,10pt,conference]{IEEEtran}

\usepackage[T1]{fontenc}          % hyphenation of accented words
\usepackage{amsmath,amssymb,amsfonts}
\usepackage{graphicx}
\usepackage{booktabs}
\usepackage{multirow}
\usepackage{pifont}
\usepackage{comment}
\usepackage{tablefootnote}
\usepackage[caption=false,font=footnotesize]{subfig}   % subfig, not subcaption:
\usepackage{cite}
\usepackage{xcolor}
\usepackage[hidelinks,breaklinks=true]{hyperref}       % load last

\usepackage{xcolor}

\definecolor{oldgray}{RGB}{140,140,140}
\definecolor{newblue}{RGB}{0,70,180}
\definecolor{commentorange}{RGB}{220,120,0}

\newif\ifrevision
\revisionfalse           % Clean final version

\ifrevision
    \newcommand{\oldtext}[1]{{\color{oldgray}#1}}
    \newcommand{\newtext}[1]{{\color{newblue}#1}}
    \newcommand{\duycomment}[1]{{\color{commentorange}#1}}
\else
    \newcommand{\oldtext}[1]{}
    \newcommand{\newtext}[1]{#1}
    \newcommand{\duycomment}[1]{}
\fi

\DeclareMathOperator{\MHA}{MHA}
\DeclareMathOperator{\LN}{LN}
\DeclareMathOperator{\MLP}{MLP}
\DeclareMathOperator{\Pool}{Pool}
\DeclareMathOperator{\PosE}{PE}

\newcommand{\cmark}{\ding{51}}

\begin{document}

\title{FINE: Future-Informed Navigation Encoding for\\ Data-Efficient Vision-Language Navigation}

\author{
  \IEEEauthorblockN{
    Khang H. Nguyen$^{1,2,*,\dagger}$\hspace{1.2em}
    Hoang Pham Quang Nguyen$^{1,*}$\hspace{1.2em}
    Ha Phuong Nguyen$^{1}$\hspace{1.2em}
    Khanh Dinh Binh$^{1}$\\[2pt]
    Xuan Ha Nguyen$^{1}$\hspace{1.2em}
    Vien Ngo$^{1,3}$\hspace{1.2em}
    Duy Ho Nguyen Minh$^{4,5,6}$\hspace{1.2em}
    Huan Nguyen$^{1,\ddagger}$\hspace{1.2em}
    An T. Le$^{1,3,7}$
  }
}

\maketitle

\renewcommand{\thefootnote}{}%
\footnotetext{
$^{*}$Equal contribution.\\
\hspace*{1em}$^{\dagger}$This work was done while the author was at VinRobotics.\\
\hspace*{1em}$^{\ddagger}$Corresponding author: \texttt{v.huannd13@vinrobotics.net}\\
\hspace*{1em}$^{1}$VinRobotics\\
\hspace*{1em}$^{2}$University of California, Los Angeles (UCLA)\\
\hspace*{1em}$^{3}$Center for AI Research, VinUniversity, Vietnam\\
\hspace*{1em}$^{4}$German Research Center for Artificial Intelligence (DFKI)\\
\hspace*{1em}$^{5}$University of Stuttgart\\
\hspace*{1em}$^{6}$Max Planck Research School for Intelligent Systems (IMPRS-IS)
\hspace*{1em}$^{7}$Intelligent Autonomous Systems, TU Darmstadt, Germany
}
\renewcommand{\thefootnote}{\arabic{footnote}}

\begin{abstract}
\oldtext{Vision-language navigation (VLN) policies built on vision-language models
generalize well but are costly to adapt: every additional environment, route and
instruction has to be demonstrated. We observe that each demonstration already
records what its action labels omit, namely which landmark the instruction refers
to, what it looks like, and how it is arranged in 3D, and that
observation-to-action training discards that signal. We introduce FINE, a
\underline{f}uture-\underline{i}nformed \underline{n}avigation \underline{e}ncoding that recovers it. FINE attaches two auxiliary
token groups to an otherwise unmodified VLN backbone: explicit landmark tokens
that regress the semantic and geometric features of the upcoming landmark region,
and an implicit token trained to separate the realized future landmark state from
world-model counterfactuals of the same scene. All supervision comes from frames
already contained in each demonstration, so FINE needs no extra human annotation, no
extra sensing and no change to the action space, and every future-dependent
module is discarded after training. On R2R-CE and RxR-CE val-unseen, FINE
adds 2.6 and 4.5 success-rate points to InternVLA-N1 at full data. Under a
reduced demonstration budget, the margin widens to 6.8 points, recovering about a
third of the success rate lost to the reduction.}
\newtext{Adapting vision-language navigation (VLN) policies to new environments is expensive because every additional route and instruction requires an embodied demonstration. Yet standard observation-to-action training uses only a small fraction of the information already contained in each trajectory. In particular, future observations reveal the instruction-relevant landmarks that the agent will encounter, including what they look like and how they are arranged in 3D. We introduce \textbf{FINE}, a \textbf{F}uture-\textbf{I}nformed \textbf{N}avigation \textbf{E}ncoding framework that extracts this latent supervision from existing demonstrations. FINE equips a VLN backbone with two complementary auxiliary representations. First, \emph{explicit landmark tokens} follow the ordered landmarks specified by the instruction and are trained to predict the future landmark region in both semantic 2D patch-feature space and viewpoint-dependent 3D geometric feature space. Second, an \emph{implicit future token} learns to distinguish the landmark state that is actually reached from plausible same-scene counterfactual futures generated by a video world model. On R2R-CE and RxR-CE val-unseen, FINE improves InternVLA-N1 by
2.6 and 4.5 success-rate points, respectively, at full training data.
More importantly, as demonstrations become limited, the benefit grows:
at a 70\% demonstration budget, FINE improves success rate by 6.8 points,
recovering roughly one-third of the performance lost by reducing the
training demonstrations. Project page is available at \textcolor{blue}{\url{https://finevln.github.io/}}.
}
\end{abstract}


\section{Introduction}
\oldtext{Vision-language navigation (VLN) requires an embodied agent to follow natural-language instructions in unseen environments from egocentric observations. Vision-language models (VLMs) have advanced the task by
transferring visual-semantic knowledge into embodied
policies~\cite{zhang2024navid,zhou2024navgpt2,cheng2024navila}. Adapting them is
expensive in a way that image-text adaptation is not: every additional
environment, route, viewpoint and instruction must be demonstrated through
sequential interaction with a 3D scene~\cite{wang2023scaling,wei2026ground}.
Reducing the number of demonstrations a VLN policy needs is therefore a practical
concern.

Most VLM-based VLN agents are trained by observation-to-action prediction,
mapping the instruction and observations directly to
a discrete action~\cite{cheng2024navila,uninavid} or a continuous
trajectory~\cite{wei2026ground}. 
However, in
Fig.~\ref{fig:realworld}, each demonstration contains richer supervision than its action labels alone.
In VLN, an instruction typically
refers to visual landmarks along
the route that may only become visible in later observations.
These future frames reveal which landmark the instruction refers to and how it is arranged spatially, which information the agent must infer at test time but an action label omits. 
Observation-to-action training discards this signal, even though it is already recorded in every trajectory and requires no extra annotation, sensing, or simulator access.
% Observation-to-action training discards it. Our starting point is that this signal is free: it requires no annotation, simulator access, or extra sensing because it is already recorded in every training trajectory.
Although future observations provide additional supervision beyond action labels, their information is not uniformly relevant to navigation.
Navigation is landmark-oriented: people follow routes using sparse reference objects, such as \emph{``walk past the sofa''} or \emph{``turn at the doorway''}, rather than memorizing entire scenes~\cite{rmus2022humans}.
% Future frames also contain instruction-irrelevant information, which can dilute supervision from relevant regions.
% We therefore restrict future supervision to the region selected by the instruction and represent it in frozen semantic and geometric feature spaces, allowing the objective to emphasize task-relevant structure over photometric detail.
Existing methods summarize what the agent has
\emph{already} seen, through observation
histories~\cite{wei2026ground,wei2025streamvln}, scene or semantic
maps~\cite{wang2021ssm}, or implicit memories~\cite{zeng2026janusvln}. 
Another predicts the future and spends that prediction \emph{at inference},
synthesizing candidate observations to score a discrete set of next
moves~\cite{bar2025navigation,zhang2026futurenav}. That works where a candidate
set exists, but it adds generative cost to every step and does not transfer to a
continuous action head. We instead uses future frames only as training supervision; all future-dependent modules are removed at inference, leaving the original policy to act directly.

We instantiate this as \textbf{FINE}, \emph{future-informed navigation encoding}:
two auxiliary token groups attached to an otherwise unmodified VLN backbone (Figure~\ref{fig:realworld}). 
The explicit branch utilizes ordered slots, one per instruction landmark, grounded in the current view and trained to predict semantic and 3D-geometric features of the corresponding future landmark region.
The implicit token is trained contrastively to distinguish the realized future landmark state from physically plausible same-scene alternatives generated by a video world model. 
Both branches supervise the region through complementary reconstructive and discriminative objectives, while a mask keeps their readouts independent within the VLM.

Because FINE changes neither the backbone architecture, action space, nor sensor suite, it can be attached to policies of different classes. We evaluate it on (i) R2R-CE under reduced demonstration budgets, on (ii) RxR-CE using the full training set, and deploy it on a (iii) humanoid robot indoors. On R2R-CE, gains are consistent and become larger as demonstrations are withheld, matching the low-data regime targeted by FINE.
We summarize our contributions as follows:
\begin{itemize}
    \item We show that the informative part of a demonstration's future is the
    instruction-selected landmark region, not the upcoming scene as a whole, and
    that supervising a VLN policy on it in frozen semantic and geometric feature
    space improves navigation without additional data.

    \item We instantiate this as FINE, a pair of attention-isolated auxiliary
    token groups: ordered landmark slots supervised by a semantic - geometric
    decoder, and an implicit token contrasted against same-scene world - model
    counterfactuals. Both are supervised by frames already present in
    each demonstration and are removed at inference.
    \item On R2R-CE under reduced demonstration budgets, FINE recovers about a third of the success rate lost to the reduction, transfers across policy classes, while also improving full-data RxR-CE performance and deploying on a humanoid robot (30 DoF, 1.72m, 75 kg) across three levels of indoor navigation difficulty.

\end{itemize}}
\duycomment{The main issues of current introductions:
\begin{itemize}
    \item The first paragraph motivates data scarcity, but the second paragraph immediately becomes descriptive. The reader does not yet get a strong “waste of supervision” argument.
Figure 1 is currently referenced as an example, but it should act as the conceptual anchor of the Introduction: instruction $\rightarrow$ ordered landmarks $\rightarrow$ future landmark evidence $\rightarrow$ explicit/implicit 
learning.
\item The paragraph beginning “Although future observations...” mixes three ideas at once: landmark relevance, prior memory methods, and future-prediction methods. These should be separated more cleanly.
\item The distinction from prior future-prediction work is good, but it currently sounds mostly like an efficiency argument. The stronger conceptual distinction is: prior work uses predicted futures to choose actions at inference; FINE uses future observations to shape representations during training.

\item The FINE paragraph is too implementation-like. “Two auxiliary token groups,” “mask,” etc. come before the reader fully internalizes why the two branches exist.

\item The evaluation paragraph is generic. It should explicitly say that the reduced-data experiment tests the central hypothesis: if FINE extracts more supervision from each demonstration, its advantage should become larger when demonstrations are scarce.
The contribution bullets can be made less architecture-oriented and more claim-oriented.

\end{itemize}
}

%%%%%%%%%%%%%%%%%%%%%%%%%%%%%%%%%%%%%%%%%%%%%%%%%%%%%%%%%%%%%%%%%%%%%%
\newtext{VLN requires an embodied agent to follow
natural-language instructions through unseen environments from egocentric
observations. Recent vision-language models (VLMs) have advanced
this task by transferring large-scale visual-semantic knowledge into embodied
navigation policies~\cite{zhang2024navid,zhou2024navgpt2,cheng2024navila}.
However, adapting these models to navigation remains expensive: \textit{every new route,
viewpoint, and instruction requires sequential interaction
with environment}~\cite{wang2023scaling,wei2026ground}. This makes the
amount of information that can be extracted from each demonstration particularly
important for data-efficient VLN.

Most VLM-based VLN policies are optimized through
\emph{observation-to-action} prediction, mapping instructions and visual observations to discrete actions~\cite{cheng2024navila,uninavid} or continuous trajectories~\cite{wei2026ground}. Yet an embodied demonstration
contains considerably richer supervision than the executed action alone.
In Fig.~\ref{fig:realworld}, the instruction specifies an
\emph{ordered sequence of landmarks}, for example, a carton box, a blue table,
an orange chair, and a green plant, that the agent should encounter as it
progresses along the route. Some landmarks are not visible from the current viewpoint and appear only in future observations. These future frames therefore provide supervision beyond the action label, revealing which landmark comes next, its appearance, and its 3D structure. This information is already contained in the demonstration and requires no additional annotation or interaction.
\begin{figure}[!t]
    \centering
    \includegraphics[width=\linewidth]{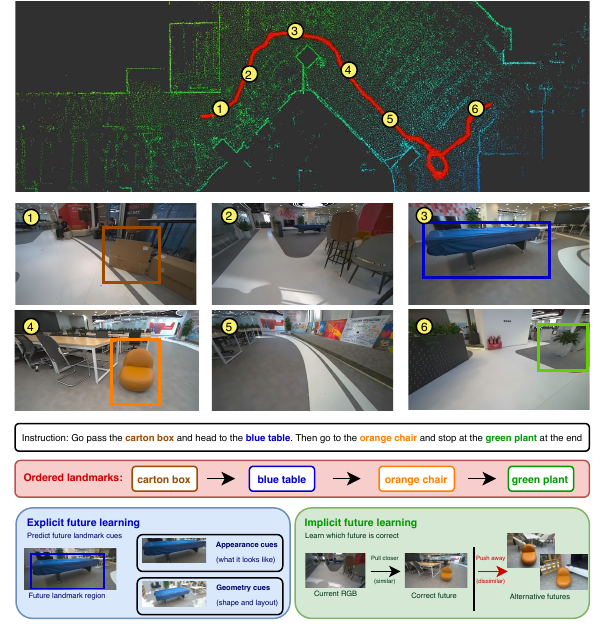}
    \vspace{-0.2in}
    \caption{
    A VLN demonstration provides supervision beyond action through future landmark observations. \textbf{Top}: a realworld rollout following carton box $\rightarrow$ blue table $\rightarrow$ orange chair $\rightarrow$ green plant; observations 2 and 5 contain no relevant landmark. \textbf{Bottom-left}: \emph{explicit future learning} predicts semantic and 3D geometric features of the upcoming landmark. \textbf{Bottom-right}: \emph{implicit future learning} distinguishes the realized future from same-scene counterfactuals.} 
    \label{fig:realworld}
    \vspace{-0.1in}
\end{figure}
%%%%%%%%%%%%%%%%%%%%%%%%%%%%%%%%%%%%%%%%%%%%%%%%%%%%%%%%%%%%%%%%%%%%%%
% A VLN demonstration contains richer supervision than action labels: future observations reveal the ordered landmarks specified by the instruction and their visual-spatial structure.
% \textbf{Top}: a real-world rollout following carton box $\rightarrow$ blue table $\rightarrow$ orange chair $\rightarrow$ green plant; observations 2 and 5 contain no instruction-relevant landmark.
% \textbf{Bottom-left}: \emph{explicit future learning} predicts semantic 2D patch features and viewpoint-dependent 3D geometric features of the upcoming landmark.
% Bottom-right: \emph{implicit future learning} distinguishes the realized future landmark state from plausible same-scene counterfactuals.}
The challenge is that future observations are not uniformly useful for navigation.
Routes are often organized around sparse visual landmarks, such as \emph{walk past the sofa} or \emph{turn at the doorway}, rather than the full scene~\cite{rmus2022humans}.
Fig.~\ref{fig:realworld} makes this
distinction explicit: observations 2 and 5 contain no instruction-relevant landmark, while the others reveal landmarks that anchor route progress. This motivates a more selective question:
rather than predicting the whole future scene, can we use the
\emph{instruction-relevant future landmark} as a training signal for the
navigation representation?

Existing VLN methods exploit temporal context in two different ways.
One line summarizes what the agent has \emph{already observed}, using observation
histories~\cite{wei2026ground,wei2025streamvln}, scene or semantic
maps~\cite{wang2021ssm}, or implicit memories~\cite{zeng2026janusvln}.
Another line predicts future observations and uses those predictions
\emph{during inference} to evaluate candidate next moves
~\cite{bar2025navigation,zhang2026futurenav}. While effective for lookahead planning, the latter adds generative computation at each navigation step and is tied to candidate-action evaluation. We take a
different view: the future should serve as a \emph{teacher during training},
rather than as an additional input or rollout mechanism at deployment.

Based on this principle, we introduce \textbf{FINE},
\emph{Future-Informed Navigation Encoding}, which extracts landmark-centered
future supervision from demonstrations through two complementary learning
signals. As shown in Fig.~\ref{fig:realworld}, the \emph{explicit} branch asks
\emph{what should appear next?} It maintains ordered landmark representations
derived from the instruction, grounds them in the current observation, and
predicts the corresponding future landmark region in both semantic 2D
patch-feature space and viewpoint-dependent 3D geometric feature space.
The \emph{implicit} branch instead asks \emph{which future is consistent with
the demonstrated route?} It learns to distinguish the realized future landmark
state from plausible same-scene counterfactual futures generated by a video
world model. The two objectives provide complementary reconstructive and discriminative supervision over the same future.

Crucially, FINE uses future information only to shape the policy representation
during training. Future frames, geometric supervision, and world-model
counterfactuals are not required as lookahead inputs at deployment, and the
policy retains its original action space and action decoder. This makes the same
future-informed learning principle applicable to different VLN policy classes,
including both discrete-action and continuous-trajectory models.

We evaluate FINE from three complementary perspectives. First, on R2R-CE, we
systematically reduce the number of available demonstrations to directly test
whether richer supervision from each trajectory becomes more valuable as data
becomes scarce. Second, we evaluate full-data generalization on both R2R-CE and
the longer, multilingual RxR-CE benchmark. Third, we deploy FINE on a real humanoid
robot under increasingly challenging landmark visibility and navigation
horizons. The results consistently support the central motivation: FINE improves
the underlying VLN policies across settings, and its advantage becomes larger
as demonstrations are withheld.}

\section{Related Work}
\label{sec:rw}

\oldtext{\noindent\textbf{Vision-Language Navigation (VLN).}
VLN has progressed from recurrent sequence models that encode instructions and navigation history for sequential action prediction~\cite{anderson2018vln} to multimodal Transformers that jointly reason over instructions, current observations, and past visual context~\cite{chen2021hamt}. More recent approaches increasingly leverage pretrained vision-language and vision-language-action models, adapting their multimodal representations to directly predict navigation actions or intermediate goals~\cite{zhang2024navid,uninavid,cheng2024navila,wei2026ground}. This shift improves semantic grounding and enables stronger transfer from large-scale pretrained models to embodied navigation. In parallel, another line of work introduces explicit spatial memory for longer-horizon reasoning. ETPNav~\cite{an2024etpnav} constructs a topological representation for planning over predicted waypoints, while MapNav~\cite{zhang2025mapnav}, JanusVLN~\cite{zeng2026janusvln}, and VLN-Zero~\cite{bhatt2025vlnzero} explore semantic, implicit-neural, and symbolic spatial representations. Together, these methods improve the agent's ability to preserve spatial context, reason over previously observed regions, and plan over long navigation horizons.}

\newtext{\noindent\textbf{Vision-Language Navigation}
VLN has evolved from recurrent and Transformer-based policies
~\cite{anderson2018vln,chen2021hamt} to pretrained VLM/VLA models that leverage
large-scale visual-language knowledge for direct action or trajectory prediction~\cite{zhang2024navid,uninavid,cheng2024navila,wei2026ground}.
In parallel, spatial-memory approaches improve long-horizon navigation by
maintaining structured information about previously observed regions, using
topological, semantic, implicit-neural, or symbolic representations
~\cite{an2024etpnav,zhang2025mapnav,zeng2026janusvln,bhatt2025vlnzero}.
These methods primarily strengthen how the agent represents its current and past
experience for grounding and planning. FINE is complementary: rather than
introducing another memory or planning representation, it exploits
\emph{future observations already contained in training demonstrations} to
supervise instruction-relevant landmark representations. To this end, FINE leverages ordered instruction landmarks to provide future-informed semantic and spatial cues during training, enriching navigation representations without additional deployment-time memory.
}

\oldtext{\noindent\textbf{Auxiliary supervision in navigation.}
Training navigation policies with objectives beyond action labels is well
established. Prior work uses auxiliary signals such as progress and orientation
prediction~\cite{ma2019selfmonitoring,ma2019regretful,zhu2020auxrn},
image-generation objectives~\cite{li2023vlnsig}, and landmark grounding
annotations~\cite{cui2023gela}. Another line predicts future observations or
features for planning~\cite{koh2021pathdreamer,zhang2026sparse,zhang2026futurenav}.
FINE differs from both lines: its targets come from frames \emph{ahead} of the
current step and are restricted to instruction-grounded landmark regions, but
the prediction modules are discarded at deployment. Thus, unlike future-prediction
methods that synthesize observations at inference to rank discrete candidates,
FINE introduces no inference-time prediction cost and applies directly to action heads. Closest to this work, FOCA~\cite{foca2026} combines
explicit prediction of task-grounded future interaction embeddings with implicit
alignment to future goal observations. We adopt this decomposition, its
token-isolation mask, and summed objective, but retarget them to navigation. 
Our explicit branch is instruction-conditioned on an \emph{ordered} set of landmarks and incorporates a geometric target provided by a frozen 3D teacher. The implicit token is obtained through query pooling, while negatives are sampled from same-scene counterfactual rollouts rather than relying solely on cross-episode features. We further leverage world-model predictions~\cite{nanowm,bar2025navigation,bagchi2026walk} to construct contrastive pairs for learning implicit future-state alignment.}
\vspace{0.05in}
\newtext{
\noindent\textbf{Future-Oriented Supervision and World Models.} Navigation policies have long benefited from objectives beyond action labels, including progress and orientation prediction ~\cite{ma2019selfmonitoring,ma2019regretful,zhu2020auxrn}, future-view generation~\cite{li2023vlnsig}, and landmark grounding ~\cite{cui2023gela}. More directly related to FINE, several works model the future for navigation. PathDreamer~\cite{koh2021pathdreamer} generates plausible future observations of unvisited viewpoints for lookahead planning, while Lookahead Exploration~\cite{wang2024hnr} predicts future environmental features to evaluate candidate paths. Navigation World Models ~\cite{bar2025navigation} and SparseVideoNav~\cite{zhang2026sparse} similarly use generated future rollouts to guide planning or trajectory selection at deployment. In contrast, FutureNav~\cite{zhang2026futurenav} incorporates future
spatial-state prediction into joint world--action modeling during training. Closest to our formulation, FOCA~\cite{foca2026} explores future-conditioned
representation learning for data-efficient VLA adaptation by leveraging both
future interaction cues and goal-state alignment. FINE instead specializes future supervision to the structure of navigation:
it uses \emph{ordered, instruction-relevant landmark regions} as the learning target, capturing both semantic 2D appearance and viewpoint-dependent 3D
geometry, without requiring future rollout for action selection at deployment.
}

\vspace{0.05in}
\oldtext{\noindent\textbf{Data-efficient Adaptation.}
Recent work has also explored data-efficient adaptation of VLN policies. He et al.~\cite{he2026jopvln} combine demonstration trajectories with reinforcement-learning exploration and evaluate few-shot adaptation to new environments. Hu et al.~\cite{hu2026adaptation} study test-time cross-domain adaptation by reusing knowledge acquired from previously encountered domains. Hong et al.~\cite{hong2025general} challenge the conventional train-once, zero-shot deployment setting and introduce scene-specific adaptation, where accumulated experience in a persistent environment is used to improve subsequent navigation. These methods change what the agent collects or how it adapts online. FINE is orthogonal, extracting more supervision from a fixed set of demonstrations without additional interaction.}

\newtext{\noindent\textbf{Data-Efficient Adaptation in VLN.}
Recent work improves VLN adaptation by acquiring or reusing experience in
different ways. JOP-VLN~\cite{he2026jopvln} combines off-policy imitation
learning and DAgger trajectories with on-policy reinforcement learning to
improve exploration and error recovery. IDEA~\cite{hu2026adaptation}
addresses test-time domain shifts by accumulating transferable adaptation
assets and reusing them across encountered domains. GSA-VLN
~\cite{hong2025general} instead studies continual adaptation to persistent
scenes, exploiting repeated experience within the same environment.
These methods improve adaptation by changing how additional experience is
collected, retained, or exploited online. FINE addresses a complementary
source of efficiency: given a \emph{fixed set of expert demonstrations}, it
extracts additional supervision from future landmark observations already
recorded in those trajectories, without requiring additional environment
interaction.}

\section{Method}
%%%%%%%%%%%%%%%%%%%%%%%%%%%%%%%%%%%%%%%%%%%%%%%%%%%%%%%%%%%%%%%%%%%%%%
\begin{figure*}[t]
    \centering
    \includegraphics[width=0.95\textwidth]{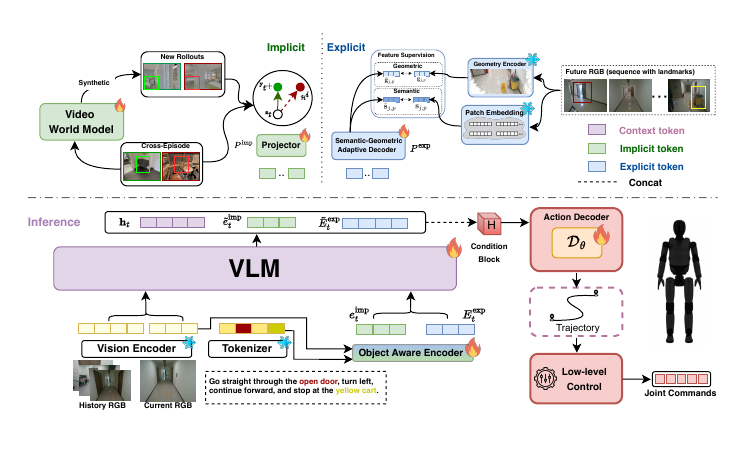}
    \vspace{-0.4in}
    \caption{
    \oldtext{\textbf{FINE architecture.}
    \emph{Bottom (train and inference):} the unmodified VLN backbone consumes the
    instruction and the history and current RGB. 
    An object-aware encoder builds $S$ explicit landmark slots
$\mathbf{E}^{\mathrm{exp}}_t$ from landmark-phrase, route-order, and empty-slot
embeddings, then grounds these latent queries in the current visual tokens
together with a shared Grounding-DINO location token
$\boldsymbol{\ell}_t$.
    A learnable query pooled
    over the frozen vision features gives the implicit token
    $\mathbf{e}^{\mathrm{imp}}_t$. Both groups are appended to the VLM context
    under a mask that blocks attention between them while preserving access to
    the shared vision--language context.
    \emph{Top (training only, discarded at inference):} 
    Within the future landmark box, SGAD predicts $\hat{\mathbf{s}}_{j,p}$ and
$\hat{\mathbf{g}}_{j,p}$ to match frozen VLM and VGGT targets
$\mathbf{s}_{j,p}$ and $\mathbf{g}_{j,p}$ under masked MSE
$\mathcal{L}_{\mathrm{exp}}$.
    A projector maps
$\tilde{\mathbf{e}}^{\mathrm{imp}}_t$ to $\mathbf{z}_t$, pulled toward the
    pooled future-landmark embedding $\mathbf{y}_{t^{+}}$ and pushed away from
    cross-episode negatives and same-scene world-model rollouts
    ($\mathcal{L}_{\mathrm{imp}}$).
    }
    \newtext{\textbf{Bottom (train and inference):} given the instruction, RGB history, and
current observation at time $t$, the object-aware encoder constructs ordered
explicit landmark tokens
$\mathbf{E}^{\mathrm{exp}}_t
=\{\mathbf{e}^{\mathrm{exp}}_{t,j}\}_{j=1}^{S}$
and an implicit future token $\mathbf{e}^{\mathrm{imp}}_t$.
After VLM processing, the resulting representations
$\widetilde{\mathbf{E}}^{\mathrm{exp}}_t$ and
$\tilde{\mathbf{e}}^{\mathrm{imp}}_t$, together with the shared context
$\mathbf{h}_t$, condition the original action decoder.
\textbf{Top (training only):} the explicit branch decodes each landmark token
into semantic predictions $\hat{\mathbf{s}}_{j,p}$ and 3D geometric predictions
$\hat{\mathbf{g}}_{j,p}$ for the corresponding future landmark region, matched
to frozen targets $\mathbf{s}_{j,p}$ and $\mathbf{g}_{j,p}$.
The implicit branch projects
$\tilde{\mathbf{e}}^{\mathrm{imp}}_t$ to $\mathbf{z}_t$, which is pulled
toward the realized future-landmark representation $\mathbf{y}_{t^{+}}$ and
pushed away from cross-episode and same-scene counterfactual futures.
\vspace{-0.2in}
}}
    \label{fig:fine_arch}
\end{figure*}
%%%%%%%%%%%%%%%%%%%%%%%%%%%%%%%%%%%%%%%%%%%%%%%%%%%%%%%%%%%%%%%%%%%%%%
\subsection{Problem Formulation and Overview}
\oldtext{\subsection{Problem Formulation and Overview}
An expert demonstration is a finite sequence
\begin{equation}
    \tau = \{(\mathbf{o}_{t},\, a_{t})\}_{t=0}^{T_{\tau}},
    \qquad a_t \in \mathcal{A},
    \label{eq:traj}
\end{equation}
of egocentric RGB observations $\mathbf{o}_{t}$ and executed actions $a_t$, paired with an instruction
$\mathcal{I}$. Given $\mathcal{I}$ and the history $\mathbf{o}_{t-T:t}$, the policy predicts
only the next action. Future observations $\mathbf{o}_{>t}$ are used purely as
training-time supervision, read from $\tau$, the policy has not access to them at test time.
The action space $\mathcal{A}$ may be discrete, as in
NaVILA~\cite{cheng2024navila}, or the space of trajectory chunks by a
diffusion policy, as in InternVLA-N1~\cite{wei2026ground}.
% Throughout, $t^{+} \triangleq \min(t+k,
% T_{\tau})$ denotes the supervising future step at offset $k$. $\mathcal{B}_{t^{+}} \subset \{1,\dots,N_p\}$ is the set of
% patch indices covered by the instruction landmark's bounding box in
% $\mathbf{o}_{t^{+}}$, used by both branches introduced below.
Throughout, \(t^{+}\triangleq \min(t+k,T_\tau)\) denotes the future supervision at offset \(k\). For landmark slot \(j\), \(\mathcal{B}_{t^{+},j}\subseteq\{1,\ldots,N_p\}\) denotes the set of patch indices covered by its bounding box in \(\mathbf{o}_{t^{+}}\), used by the branches introduced below.

FINE is a plug-in for any VLN backbone $\pi_\theta$ that maps
$(\mathcal{I}, \mathbf{o}_{t-T:t})$ to a hidden representation
$\mathbf{h}_t$ for downstream action prediction. An overview is shown in Fig.\ref{fig:fine_arch}. It augments that representation
with a set of $S$ explicit landmark tokens
$\mathbf{E}^{\mathrm{exp}}_t = [\mathbf{e}^{\mathrm{exp}}_{t,1},\dots,
\mathbf{e}^{\mathrm{exp}}_{t,S}]^{\!\top} \in \mathbb{R}^{S \times H}$ and a
single implicit token $\mathbf{e}^{\mathrm{imp}}_t \in \mathbb{R}^{H}$
(Fig.~\ref{fig:fine_arch}), which are appended to the multimodal context and
processed jointly by the VLM:
\begin{equation}
    \mathbf{h}_t,\ \tilde{\mathbf{E}}^{\mathrm{exp}}_t,\ \tilde{\mathbf{e}}^{\mathrm{imp}}_t
    = \pi_\theta\bigl(\mathcal{I},\, \mathbf{o}_{t-T:t},\,
      \mathbf{E}^{\mathrm{exp}}_t,\, \mathbf{e}^{\mathrm{imp}}_t\bigr).
    \label{eq:fine_forward}
\end{equation}
The action is then produced by the backbone's own head $\mathcal{D}_\theta$,
unchanged:
\begin{equation}
    a_t \sim \mathcal{D}_\theta\bigl(\cdot \mid \mathbf{h}_t,\,
      \tilde{\mathbf{E}}^{\mathrm{exp}}_t,\, \tilde{\mathbf{e}}^{\mathrm{imp}}_t\bigr),
    \qquad a_t \in \mathcal{A}.
    \label{eq:fine_action}
\end{equation}
The auxiliary-token formulation, the token-isolation mask of
Sec.~\ref{sec:policy} and the summed objective of Eq.~\eqref{eq:total} follow
FOCA~\cite{foca2026}, which introduced them for manipulation. Sec.~\ref{sec:rw}
states what navigation changes.}

\newtext{
\label{sec:overview}
As shown in Fig.~\ref{fig:fine_arch}, FINE uses future observations already
contained in an expert demonstration as additional training supervision.
Given an instruction $\mathcal{I}$, a demonstration is
\begin{equation}
    \tau=\{(\mathbf{o}_t,a_t)\}_{t=0}^{T_\tau},
    \qquad a_t\in\mathcal{A},
\end{equation}
where $\mathbf{o}_t$ is the egocentric RGB observation and $a_t$ the executed
action. At time $t$, the policy observes only
$(\mathcal{I},\mathbf{o}_{t-T:t})$; future observations are available only
during training.

Rather than supervising the policy on the entire future scene, FINE focuses on
the instruction-relevant landmark that appears later in the demonstration.
Let
\(
t^+=\min(t+k,T_\tau)
\)
denote the future supervision step and
$\mathcal{B}_{t^+,j}$ the patch region corresponding to landmark $j$.
FINE extracts two complementary learning signals from this region:
(i) \emph{landmark prediction}, which learns what semantic and geometric
evidence should appear next, and
(ii) \emph{future-state discrimination}, which learns which plausible future
is consistent with the demonstrated route.
For a VLN backbone $\pi_\theta$, FINE introduces $S$ ordered landmark tokens
\[
\mathbf{E}^{\mathrm{exp}}_t
=
[\mathbf{e}^{\mathrm{exp}}_{t,1},\ldots,
 \mathbf{e}^{\mathrm{exp}}_{t,S}]^\top
\]
and one future-state token $\mathbf{e}^{\mathrm{imp}}_t$.
They are processed jointly with the original multimodal context:
\begin{equation}
(\mathbf{h}_t,
 \widetilde{\mathbf{E}}^{\mathrm{exp}}_t,
 \widetilde{\mathbf{e}}^{\mathrm{imp}}_t)
=
\pi_\theta(
\mathcal{I},
\mathbf{o}_{t-T:t},
\mathbf{E}^{\mathrm{exp}}_t,
\mathbf{e}^{\mathrm{imp}}_t).
\label{eq:fine_forward}
\end{equation}
The resulting representations condition the backbone's original action head,
so FINE applies to both discrete-action and continuous-trajectory policies.}

\oldtext{\subsection{Explicit Landmark Tokens}
\label{sec:explicit}
VLMs carry strong object semantics but are comparatively weak at the
viewpoint-dependent spatial structure a moving agent needs.
The explicit branch maintains ordered landmark representations, grounds them in the current view, and predicts their future appearance and 3D geometry.

\subsubsection{Slot initialization}
Explicit slots are latent queries, with each slot reserved for one landmark along the instructed route.
They initially encode landmark identity and order in the instruction, then ground in the current observation to form explicit tokens that predict future semantic and geometric cues.
In the Object Aware Encoder, we first parse $\mathcal{I}$ once per episode into an ordered list of at most $M$
landmark phrases and allocate $S$ slots. Slot $j$ is initialized as
\begin{equation}
    \mathbf{q}^{\mathrm{exp}}_j
    = \mu_j\,\mathbf{c}_j + \mathbf{r}_j + (1-\mu_j)\,\mathbf{p},
    \qquad j = 1,\dots,S,
    \label{eq:slots}
\end{equation}
where $\mathbf{c}_j \in \mathbb{R}^{H}$ is the frozen text-encoder embedding of
the $j$-th landmark phrase, $\mathbf{r}_j$ a learned embedding of its rank along
the route, $\mathbf{p}$ a shared learnable empty-slot token, and
$\mu_j = \mathbb{1}[\,j \le \min(M,S)\,]$ marks slot validity. When $M > S$ we
keep the first $S$ landmarks in order of mention.
We extract the ordered landmark list offline, once per instruction, using a fixed LLM prompt that returns concise noun phrases in mention order. Then, Grounding DINO~\cite{groundingdino} is used only for localizing objects in RGB frames and not for parsing $\mathcal{I}$.

At this stage, she slots encode which landmarks to track but are not yet conditioned on the current scene.
The slot queries $\mathbf{Q} = [\mathbf{q}^{\mathrm{exp}}_1,\dots,
\mathbf{q}^{\mathrm{exp}}_S]^{\!\top}$ are 
grounded in the current observation by
cross-attention:
\begin{equation}
\begin{aligned}
\mathbf{G} &= \mathbf{Q} + \MHA\bigl(\mathbf{Q},\,
    \LN([\mathbf{V}_t;\boldsymbol{\ell}_t]),\, \LN([\mathbf{V}_t;\boldsymbol{\ell}_t])\bigr),\\
\mathbf{E}^{\mathrm{exp}}_t &= \mathbf{G} + \MLP\bigl(\LN(\mathbf{G})\bigr),
\end{aligned}
\label{eq:explicit_init}
\end{equation}
where $[\,\cdot\,;\,\cdot\,]$ concatenates along the token axis,
$\mathbf{V}_t \in \mathbb{R}^{N_v \times H}$ are the current visual tokens, and
$\boldsymbol{\ell}_t = \MLP_{\mathrm{loc}}(\PosE(\mathcal{B}_t))
\in \mathbb{R}^{1 \times H}$ encodes the object-of-interest box detected in
$\mathbf{o}_t$ by Grounding DINO; all matched detections in a frame are included in the mask.
$\boldsymbol{\ell}_t$ is one shared OOI-mask token, and learned queries are not
pre-normalized. Each explicit token therefore carries landmark identity, its
grounding in the current view, and its position along the route.
% TODO v1 review:\revq{state whether $\boldsymbol{\ell}_t$ is one shared location token or one per landmark.}-> DONE

\subsubsection{Semantic--geometric Supervision}
A Semantic--Geometric Adaptive Decoder (SGAD)  maps each processed token
$\tilde{\mathbf{e}}^{\mathrm{exp}}_{t,j} \in \tilde{\mathbf{E}}^{\mathrm{exp}}_t$ to a per-patch prediction of the
landmark's future region. SGAD is a lightweight pre-norm Transformer decoder ($\approx 45.6$M parameters), spatially conditioned on a sinusoidal encoding of the
landmark box.
% TODO v1 review -> DONE
% \revq{give SGAD's depth, width, head count and parameter budget, and replace
% ``channel-wise scaling'' with the setting used in the reported runs.}
Targets are extracted from $\mathbf{o}_{t^{+}}$ by two frozen teachers: the
backbone's own patch embedding $f_{\mathrm{enc}}$ for semantics, and
VGGT Geometry Encoder~\cite{vggt} as $f_{\mathrm{geo}}$ for viewpoint-dependent 3D structure.
Using the future-region patch set $\mathcal{B}_{t^{+},j}$,
% Writing $\mathcal{B}_j$ for the patches covered by slot $j$'s landmark box in
% $\mathbf{o}_{t^{+}}$, 
the explicit loss is a masked MSE, normalized per valid
slot:
\begin{equation}
\begin{aligned}
\mathcal{L}_{\mathrm{exp}}
= \frac{1}{\sum_{j}\mu_j}\sum_{j=1}^{S} \frac{\mu_j}{|\mathcal{B}_{t^{+},j}|}
  \sum_{p\in\mathcal{B}_{t^{+}}} \Bigl(\;
  &\bigl\|\hat{\mathbf{s}}_{j,p}-\mathbf{s}_{j,p}\bigr\|_2^{2} \\[-8pt]
  &+\; \beta\,\bigl\|\hat{\mathbf{g}}_{j,p}-\mathbf{g}_{j,p}\bigr\|_2^{2} \Bigr),
\end{aligned}
\label{eq:lexp}
\end{equation}
with $\mathbf{s}_{j,p} = f_{\mathrm{enc}}(\mathbf{o}_{t^{+}})_p$,
$\mathbf{g}_{j,p} = f_{\mathrm{geo}}(\mathbf{o}_{t^{+}})_p$, 
where $p$ indexes patches within the future landmark region
$\mathcal{B}_{t^{+},j}$.
and $\beta$
balancing two terms that live in different feature spaces. Slots with
$\mu_j = 0$ or $\mathcal{B}_{t^{+},j} = \emptyset$ are excluded from both sums and
contribute no gradient. 
% The second case occurs whenever the landmark is not
% visible at $t^{+}$, which partial observability makes common.
We use the \(k\)-offset future frame \(t^{+}\) defined above (\(k=10\)); if the landmark is not visible, we use its nearest future visible frame.}

% TODO v1 review -> DONE
% \revq{report the skip rate for invisible landmarks, and state whether $k$ is a
% fixed offset or a maximum searched for the nearest frame in which the landmark is
% visible.}

\newtext{\subsection{Ordered Landmark Prediction}
\label{sec:explicit}

Navigation instructions typically describe a sequence of reference objects
rather than a single goal. For example, an agent may be instructed to
\emph{pass the carton box}, then \emph{head toward the blue table}, and finally
\emph{stop at the green plant}. We therefore represent instruction-relevant
landmarks as \emph{ordered slots}, such that each representation captures both
\emph{what} landmark should be encountered and \emph{where it occurs} along the
instructed route. These landmark representations are first grounded in the
current observation and then supervised to predict the semantic appearance and
3D structure of their corresponding future regions.

\paragraph{Ordered landmark token construction}
We parse the instruction $\mathcal{I}$ once per episode into an ordered list of
at most $M$ landmark phrases and allocate $S$ landmark slots.
For the $j$-th slot, we initialize a latent query as
\begin{equation}
    \mathbf{q}^{\mathrm{exp}}_j
    =
    \mu_j\,\mathbf{c}_j
    +
    \mathbf{r}_j
    +
    (1-\mu_j)\,\mathbf{p},
    \qquad j=1,\ldots,S,
    \label{eq:slots}
\end{equation}
where $\mathbf{c}_j \in \mathbb{R}^{H}$ is the frozen text embedding of the
$j$-th landmark phrase, $\mathbf{r}_j$ is a learned embedding of its order
along the route, and $\mathbf{p}$ is a shared learnable empty-slot embedding.
The indicator
\(
\mu_j=\mathbb{1}[j\leq\min(M,S)]
\)
marks whether slot $j$ corresponds to a valid landmark.
When $M>S$, we retain the first $S$ landmarks in their instruction order.

The ordered landmark list is extracted offline using a fixed LLM prompt that
returns concise landmark phrases in mention order. At each observation,
Grounding DINO~\cite{groundingdino} localizes the corresponding objects and
provides a location cue $\boldsymbol{\ell}_t$.
The landmark queries
\[
    \mathbf{Q}
    =
    [\mathbf{q}^{\mathrm{exp}}_1,\ldots,
     \mathbf{q}^{\mathrm{exp}}_S]^{\top}
\]
are then grounded in the current visual tokens $\mathbf{V}_t$ through the
object-aware encoder (OAE):
\begin{equation}
    \mathbf{E}^{\mathrm{exp}}_t
    =
    \mathrm{OAE}
    \bigl(
        \mathbf{Q},
        \mathbf{V}_t,
        \boldsymbol{\ell}_t
    \bigr),
    \label{eq:explicit_init}
\end{equation}
where
\(
\mathbf{E}^{\mathrm{exp}}_t
=
[\mathbf{e}^{\mathrm{exp}}_{t,1},\ldots,
 \mathbf{e}^{\mathrm{exp}}_{t,S}]^{\top}
\)
contains the grounded landmark tokens.
$\mathrm{OAE}$ is implemented by cross-attention from the landmark queries to
the current visual tokens and location cue, followed by a feed-forward block.
Each token therefore combines landmark identity, route order, and evidence from
the current view.

\paragraph{Semantic--geometric future supervision}
Knowing which landmark should be encountered is not sufficient for navigation:
the representation should also anticipate \emph{what it will look like} and
\emph{how it will be spatially arranged} from a future viewpoint. We therefore
supervise each processed landmark token
$\widetilde{\mathbf{e}}^{\mathrm{exp}}_{t,j}$
using two complementary feature spaces.

For patches
$p\in\mathcal{B}_{t^{+},j}$
within the corresponding future landmark region, we extract frozen target
features
\begin{equation}
    \mathbf{s}_{j,p}
    =
    f_{\mathrm{enc}}(\mathbf{o}_{t^{+}})_p,
    \qquad
    \mathbf{g}_{j,p}
    =
    f_{\mathrm{geo}}(\mathbf{o}_{t^{+}})_p,
    \label{eq:future_targets}
\end{equation}
where $f_{\mathrm{enc}}$ denotes the backbone's frozen visual patch encoder and
$f_{\mathrm{geo}}$ is the frozen VGGT geometry encoder~\cite{vggt}.
The semantic feature $\mathbf{s}_{j,p}$ captures the visual identity and
appearance, while $\mathbf{g}_{j,p}$ captures
viewpoint-dependent 3D structure.

A Semantic--Geometric Adaptive Decoder (SGAD) maps
$\widetilde{\mathbf{e}}^{\mathrm{exp}}_{t,j}$
to corresponding predictions
$\hat{\mathbf{s}}_{j,p}$ and
$\hat{\mathbf{g}}_{j,p}$.
Let
\begin{equation}
    \mathcal{V}_t
    =
    \left\{
        j\; \middle|\;
        \mu_j=1,\;
        \mathcal{B}_{t^{+},j}\neq\emptyset
    \right\}
    \label{eq:valid_slots}
\end{equation}
denote the set of valid landmark slots with available future supervision.
The explicit landmark objective is
\begin{equation}
\begin{aligned}
    \mathcal{L}_{\mathrm{exp}}
    =
    \frac{1}{|\mathcal{V}_t|}
    \sum_{j\in\mathcal{V}_t}
    \frac{1}{|\mathcal{B}_{t^{+},j}|}
    \sum_{p\in\mathcal{B}_{t^{+},j}}
    \Big(
        &\left\|
            \hat{\mathbf{s}}_{j,p}
            -
            \mathbf{s}_{j,p}
        \right\|_2^2
        \\
        &+
        \beta
        \left\|
            \hat{\mathbf{g}}_{j,p}
            -
            \mathbf{g}_{j,p}
        \right\|_2^2
    \Big),
\end{aligned}
\label{eq:lexp}
\end{equation}
where $\beta$ balances the semantic and geometric feature spaces.

We use the future step $t^{+}$ defined in
Sec.~\ref{sec:overview} with offset $k=10$.
If landmark $j$ is not visible at that step, we use its nearest subsequent
visible observation.
Thus, rather than reconstructing the entire future scene,
$\mathcal{L}_{\mathrm{exp}}$ concentrates supervision on the
\emph{instruction-relevant region}, teaching each ordered landmark token what
semantic and geometric evidence should emerge as the agent progresses along the
route.}
\oldtext{\subsection{Implicit Future-State Contrast}
\label{sec:implicit}

While the explicit branch performs reconstruction, recovering the appearance of the upcoming landmark, the implicit branch performs discrimination on that same region. It learns to distinguish the future the agent actually reaches from futures it could plausibly have reached instead. Both branch attend to the same region and complement each other in kind.

An implicit token is initialized by a learnable query attending over the frozen
feature map $\mathbf{X}_t = f_{\mathrm{enc}}(\mathbf{o}_t)$, in place of mean
pooling, so that spatial disparity is preserved:
\begin{equation}
\begin{aligned}
    \mathbf{u}_t &= \mathbf{q}^{\mathrm{imp}} +
        \MHA\bigl(\mathbf{q}^{\mathrm{imp}},\, \LN(\mathbf{X}_t),\, \LN(\mathbf{X}_t)\bigr),\\
    \mathbf{e}^{\mathrm{imp}}_t &= \mathbf{u}_t + \MLP\bigl(\LN(\mathbf{u}_t)\bigr).
\end{aligned}
\label{eq:implicit_init}
\end{equation}
The index $i$ enumerates anchors in a minibatch $\mathcal{N}_{\mathrm{b}}$; anchor $i$ is the pair (episode $m(i)$, step $t(i)$). After VLM processing, a learned head $P^{\mathrm{imp}}$
projects the token into a $d_e$-dimensional contrastive space, where
its positive is the frozen-encoder embedding pooled over the future landmark
region:
\begin{equation}
    \mathbf{z}^{i}_t = P^{\mathrm{imp}}\bigl(\tilde{\mathbf{e}}^{\mathrm{imp}}_t\bigr)
      \in \mathbb{R}^{d_e},
    \quad
    \mathbf{y}^{i}_{t^{+}} = \Pool\bigl(f_{\mathrm{enc}}(\mathbf{o}^{i}_{t^{+}})
      % \mid \mathcal{B}^{i}_{t^{+}}\bigr).
      \mid \mathcal{B}^{i}_{t^{+},j}\bigr).
    \label{eq:proj_pos}
\end{equation}

\includecomment{
\noindent\textbf{Negatives.}
Negatives drawn from other episodes rarely test route-level discrimination:
two episodes even sharing a scene is already unlikely given the dataset's
environment diversity, and conditioned on sharing one, the paired segment is
more likely a distant, easily separable part of that scene than a genuinely
informative negative. Such a token can satisfy the objective by recognizing
the scene alone, without representing where in the route the agent is. We
therefore add same-scene negatives obtained by rolling out a frozen video
world model $W$ from $\mathbf{o}_t$ under an action sequence whose terminal
actions are perturbed, leaving the earlier prefix intact so that early scene
content stays shared. Let $t_{\mathrm{div}}$ be the first step at which the
perturbed sequence departs from the demonstrated one. The negative set for
anchor $i$ in training phase $\kappa$ is
\begin{equation}
\begin{aligned}
\mathcal{H}^{i}_{\kappa} =\;
&\underbrace{\bigl\{\, \mathbf{y}^{i'}_{t^{+}} \;:\; i'\in\mathcal{N}_{\mathrm{b}},\
   m(i')\neq m(i) \,\bigr\}}_{\text{cross-episode}} \\
\cup\;
&\underbrace{\bigl\{\, \Pool\bigl(f_{\mathrm{enc}}(\hat{\mathbf{o}}^{\,i}_{\tilde t})
   \mid \hat{\mathcal{B}}^{\,i}_{{\tilde t},j}\bigr) \;:\;
   \tilde t > \max\bigl(t,\, t_{\mathrm{div}}\bigr) \,\bigr\}}_{\text{world-model rollout, phase 2 only}}
\end{aligned}
\label{eq:negatives}
\end{equation}
where $\hat{\mathbf{o}}^{\,i}_{\tilde t}$ are frames of the rollout and
$\hat{\mathcal{B}}^{\,i}_{{\tilde t},j}$ the landmark box re-detected on them.
Anchors whose landmark is not detected contribute no rollout negative. We perturb only the trailing $k = 2$ actions of the $H = 10$-step rollout, so the
prefix up to $t_{\mathrm{div}} = t + H - k + 1$ matches the demonstration
exactly; frames generated before $t_{\mathrm{div}}$ are therefore near-copies
of the demonstrated future. The bound $\tilde t > \max(t, t_{\mathrm{div}})$
excludes these frames from the negative set, since admitting them would push
the anchor away from content that its positive simultaneously pulls it toward.

% \rev{The
% bound $\tilde t > \max(t, t_{\mathrm{div}})$ is required: because only terminal
% actions are perturbed, every generated frame \emph{before} $t_{\mathrm{div}}$ is
% a near-copy of the demonstrated future, and admitting it would push the anchor
% away from content the positive simultaneously pulls it toward.}
% \revq{state how many trailing actions are perturbed, which fixes
% $t_{\mathrm{div}}$, and the relation between $k$ and $t_{\mathrm{div}}$.}

The token is trained with InfoNCE~\cite{oord2018cpc} over the positive and
$\mathcal{H}^{i}_{\kappa}$:
\begin{equation}
\mathcal{L}_{\mathrm{imp}}
= -\frac{1}{|\mathcal{N}_{\mathrm{b}}|}\sum_{i\in\mathcal{N}_{\mathrm{b}}}
  \log \frac{\exp\bigl(s(\mathbf{z}^{i}_t,\, \mathbf{y}^{i}_{t^{+}})/\lambda\bigr)}
            {\displaystyle\sum_{\mathbf{y}\in\{\mathbf{y}^{i}_{t^{+}}\}\cup\mathcal{H}^{i}_{\kappa}}
             \exp\bigl(s(\mathbf{z}^{i}_t,\, \mathbf{y})/\lambda\bigr)},
\label{eq:infonce}
\end{equation}
with $s(\cdot,\cdot)$ the cosine similarity and $\lambda$ a temperature. Anchor
indexing is required: a single shared negative set would let an anchor's own
positive act as its own negative.}

\noindent\textbf{Negatives.}
Negatives drawn from other episodes are usually too easy to force route-level discrimination. Most sampled episodes occur in a different scene altogether, given the dataset's environment diversity. Then the token can satisfy the objective by recognizing the scene alone, without representing the agent's position along the route. Even when two episodes share a scene, the paired segment is usually a spatially distant part of it, and trivially easy to discriminate. It rarely acts as a hard negative - the kind that would force the token to distinguish a left turn from a right. We therefore add same-scene negatives obtained by rolling out a frozen video world model $W$ from $\mathbf{o}_t$ under an action sequence whose terminal actions are perturbed, leaving the earlier prefix intact so that early scene content stays shared. Let $t_{\mathrm{div}}$ be the first step at which the perturbed sequence departs from the demonstrated one. The negative set for
anchor $i$ in training phase $\kappa$ is
\begin{equation}
\begin{aligned}
\mathcal{H}^{i}_{\kappa} =\;
&\underbrace{\bigl\{\, \mathbf{y}^{i'}_{t^{+}} \;:\; i'\in\mathcal{N}_{\mathrm{b}},\
   m(i')\neq m(i) \,\bigr\}}_{\text{cross-episode}} \\
\cup\;
&\underbrace{\bigl\{\, \Pool\bigl(f_{\mathrm{enc}}(\hat{\mathbf{o}}^{\,i}_{\tilde t})
   \mid \hat{\mathcal{B}}^{\,i}_{{\tilde t},j}\bigr) \;:\;
   \tilde t > \max\bigl(t,\, t_{\mathrm{div}}\bigr) \,\bigr\}}_{\text{world-model rollout, phase 2 only}}
\end{aligned}
\label{eq:negatives}
\end{equation}
where $\hat{\mathbf{o}}^{\,i}_{\tilde t}$ are frames of the rollout and
$\hat{\mathcal{B}}^{\,i}_{{\tilde t},j}$ the landmark box re-detected on them.
Anchors whose landmark is not detected contribute no rollout negative. We perturb only the trailing $k = 2$ actions of the $H = 10$-step rollout, so the
prefix up to $t_{\mathrm{div}} = t + H - k + 1$ matches the demonstration
exactly; frames generated before $t_{\mathrm{div}}$ are therefore near-copies
of the demonstrated future. The bound $\tilde t > \max(t, t_{\mathrm{div}})$
excludes these frames from the negative set, since admitting them would push
the anchor away from content that its positive simultaneously pulls it toward.

}

\newtext{\subsection{Future-State Discrimination}
\label{sec:implicit}

Ordered landmark prediction teaches the representation \emph{what should appear
next}, but reconstruction alone does not require the policy to distinguish the
demonstrated future from another visually plausible outcome. For navigation,
this distinction is important: two futures may share similar scene content while
corresponding to different route progress. We therefore introduce a
complementary discriminative objective that asks:
\emph{which future is consistent with the demonstrated route?}

\paragraph{Future-state representation.}
Rather than assigning one token to each landmark, we summarize the current visual
state with a single future-state token. Given the frozen visual feature map
\(
\mathbf{X}_t=f_{\mathrm{enc}}(\mathbf{o}_t)
\),
a learnable query aggregates spatially relevant evidence through query pooling:
\begin{equation}
    \mathbf{e}^{\mathrm{imp}}_t
    =
    \mathrm{QPool}
    \bigl(
        \mathbf{q}^{\mathrm{imp}},
        \mathbf{X}_t
    \bigr),
    \label{eq:implicit_init}
\end{equation}
where $\mathrm{QPool}$ is implemented with cross-attention followed by a
feed-forward block. Unlike mean pooling, the learnable query can selectively
attend to spatially informative regions of the current observation.

After VLM processing, the token
$\widetilde{\mathbf{e}}^{\mathrm{imp}}_t$
is projected into a contrastive embedding space:
\begin{equation}
    \mathbf{z}^{i}_t
    =
    P^{\mathrm{imp}}
    \bigl(
        \widetilde{\mathbf{e}}^{\mathrm{imp},i}_t
    \bigr),
    \label{eq:implicit_proj}
\end{equation}
where $i$ indexes an anchor in minibatch $\mathcal{N}_{\mathrm{b}}$.
Its positive target is the frozen visual representation pooled over the
corresponding future landmark region:
\begin{equation}
    \mathbf{y}^{i}_{t^{+}}
    =
    \Pool
    \left(
        f_{\mathrm{enc}}(\mathbf{o}^{i}_{t^{+}})
        \mid
        \mathcal{B}^{i}_{t^{+},j(i)}
    \right),
    \label{eq:implicit_positive}
\end{equation}
where $j(i)$ denotes the landmark associated with the future supervision of
anchor $i$. The objective therefore aligns the current representation with the
landmark state that the demonstrated trajectory eventually reaches.

\paragraph{Same-scene counterfactuals.}
Cross-episode negatives are often too easy for route-level discrimination:
the model can distinguish them using scene identity without learning where the
agent lies along the instructed route. To construct harder negatives, we roll
out a frozen video world model $W$ from the current observation
$\mathbf{o}_t$ while perturbing the terminal part of the demonstrated action
sequence. The generated trajectory therefore preserves much of the same visual
context initially, but eventually diverges toward a different future state.
These \emph{same-scene counterfactuals} encourage the representation to capture
route progress rather than scene identity alone.

For anchor $i$, we denote the negative set as
\begin{equation}
    \mathcal{H}^{i}_{\kappa}
    =
    \mathcal{H}^{i}_{\mathrm{cross}}
    \cup
    \mathcal{H}^{i}_{\mathrm{cf}}(\kappa),
    \label{eq:negative_sets}
\end{equation}
where $\mathcal{H}^{i}_{\mathrm{cross}}$ contains future-landmark features from
other episodes, while $\mathcal{H}^{i}_{\mathrm{cf}}$ contains landmark
features extracted from same-scene world-model rollouts after they diverge from
the demonstrated trajectory. Counterfactual negatives are introduced only in
the second training phase.

The future-state token is then trained with InfoNCE:
\begin{equation}
\begin{aligned}
\mathcal{L}_{\mathrm{imp}}
&=
-\frac{1}{|\mathcal{N}_{\mathrm{b}}|}
\sum_{i\in\mathcal{N}_{\mathrm{b}}}
\log
\frac{
    \exp\!\left(
        s(\mathbf{z}^{i}_t,\mathbf{y}^{i}_{t^{+}})/\lambda
    \right)
}{
    Z_i
},\\
Z_i
&=
\exp\!\left(
    s(\mathbf{z}^{i}_t,\mathbf{y}^{i}_{t^{+}})/\lambda
\right)
+
\sum_{\mathbf{y}\in\mathcal{H}^{i}_{\kappa}}
\exp\!\left(
    s(\mathbf{z}^{i}_t,\mathbf{y})/\lambda
\right).
\end{aligned}
\label{eq:infonce}
\end{equation}
where $s(\cdot,\cdot)$ denotes cosine similarity and $\lambda$ is the
temperature. This objective pulls the current representation toward the demonstrated future landmark and away from plausible same-scene alternatives.
}

\oldtext{\subsection{Token Isolation and Training}
\label{sec:policy}

\noindent\textbf{Token isolation.}
Both branches supervise the same future region, allowing the implicit token to potentially exploit information encoded by explicit tokens. We prevent this by blocking implicit-to-explicit attention throughout the VLM. For the packed sequence \([\mathbf{x}_t;\mathbf{E}^{\mathrm{exp}}_t;\mathbf{e}^{\mathrm{imp}}_t]\), the corresponding entries of \(M\) are set to \(-\infty\):
$
\mathrm{Attn}(\mathbf{Q},\mathbf{K},\mathbf{V})
=\mathrm{softmax}(\mathbf{Q}\mathbf{K}^{\!\top}/\sqrt{d}+M)\mathbf{V}.
$
The implicit token must therefore derive its representation from the shared multimodal context, while explicit tokens are processed by SGAD. The backbone's causal mask already blocks explicit-to-implicit attention, while both branches retain access to \(\mathbf{x}_t\). Thus, the mask enforces readout independence, with complementarity induced by their distinct objectives.

\noindent\textbf{Objective.}
FINE preserves the backbone's original policy loss, token-level cross-entropy for
discrete actions or a diffusion or flow-matching objective for continuous ones,
and adds the two auxiliary terms under a two-phase schedule:
\begin{equation}
\mathcal{L}_{\mathrm{total}}(\kappa)
= w_a(\kappa)\,\mathcal{L}_{\mathrm{action}}
+ w_e(\kappa)\,\mathcal{L}_{\mathrm{exp}}
+ w_i(\kappa)\,\mathcal{L}_{\mathrm{imp}}\bigl(\mathcal{H}_{\kappa}\bigr),
\label{eq:total}
\end{equation}
where $\kappa \in \{1,2\}$ indexes the phase. Phase~1 emphasizes representation
learning and restricts $\mathcal{H}_1$ to cross-environment negatives; phase~2
emphasizes the policy and adds the world-model negatives of
Eq.~\eqref{eq:negatives}. Baselines use $w_e=w_i=0$ with the same step budget. Weights are in Sec.~\ref{sec:impl}.

\noindent\textbf{Inference.}
At test time, we discard Grounding DINO, SGAD, VGGT, the implicit projector \(P^{\mathrm{imp}}\), and the world model \(W\). FINE retains only the instruction landmark parse, the token initializers in Eqs.~\eqref{eq:slots}--\eqref{eq:implicit_init}, and \(S{+}1\) additional VLM tokens, adding \(\approx143\)M parameters (\(\approx1.7\%\) of the 8B backbone). Inference requires neither future observations nor generative rollouts, distinguishing FINE from per-step lookahead methods and allowing direct use with continuous-action policies.
}

\newtext{\subsection{Training and Inference}
\label{sec:policy}

\paragraph{Training}
The two future objectives supervise the same landmark region but capture
complementary information. To prevent one representation from directly
exploiting the other, we {block attention between the explicit landmark tokens
and the implicit future token}, while allowing both to attend to the shared
vision-language context.

FINE preserves the original VLN action objective and augments it with the two
future-supervision losses:
\begin{equation}
    \mathcal{L}_{\mathrm{total}}
    =
    w_a\,\mathcal{L}_{\mathrm{action}}
    +
    w_e\,\mathcal{L}_{\mathrm{exp}}
    +
    w_i\,\mathcal{L}_{\mathrm{imp}},
    \label{eq:total}
\end{equation}
where $\mathcal{L}_{\mathrm{exp}}$ learns what semantic and geometric landmark
evidence should appear next, while $\mathcal{L}_{\mathrm{imp}}$ learns which
future is consistent with the demonstrated route. Training follows a two-stage
schedule detailed in Sec.~\ref{sec:impl}.

\paragraph{Inference.}
Future observations and supervision modules are training-only. At deployment, the policy acts from the instruction and observed RGB sequence, without future generation or world-model rollout. FINE therefore introduces no generative lookahead for
action selection and applies to both discrete-action and continuous-trajectory
VLN policies.}

\section{Experiments}
%%%%%%%%%%%%%%%%%%%%%%%%%%%%%%%%%%%%%%%%%%%%%%%%%%%%%%%%%%%%%%%%%%%%%%
\begin{table}[t]
\centering
\caption{Demonstration-budget comparison on R2R-CE val-unseen.
Baseline and FINE use identical training subsets and matched optimization
within each budget. Differences are in points.}
\label{tab:dataeff}
\footnotesize\setlength{\tabcolsep}{4pt}\renewcommand{\arraystretch}{1.05}
\vspace{-0.05in}
\begin{tabular}{llcccc}
\toprule
\textbf{Budget} & \textbf{Method}
& \textbf{NE}$\downarrow$ & \textbf{OS}$\uparrow$
& \textbf{SR}$\uparrow$ & \textbf{SPL}$\uparrow$ \\
\midrule
\multirow{4}{*}{100\%}
 & NaVILA \cite{cheng2024navila}       & 5.20 & 61.6 & 54.0 & 49.0 \\
 & \;\;$+$ FINE                        & \textbf{5.00} & \textbf{63.1} & \textbf{55.6} & \textbf{50.7} \\
 \cmidrule(lr){2-6}
 & InternVLA-N1 \cite{wei2026ground}   & 4.83 & 63.3 & 58.2 & 54.0 \\
 & \;\;$+$ FINE                        & \textbf{4.29} & \textbf{65.7} & \textbf{60.8} & \textbf{55.2} \\
\midrule
\multirow{4}{*}{70\%}
 & NaVILA \cite{cheng2024navila}       & 6.76 & 46.9 & 36.2 & 29.5 \\
 & \;\;$+$ FINE                        & \textbf{6.19} & \textbf{50.6} & \textbf{41.7} & \textbf{35.5} \\
 \cmidrule(lr){2-6}
 & InternVLA-N1 \cite{wei2026ground}   & 6.45 & 46.0 & 38.5 & 34.4 \\
 & \;\;$+$ FINE                        & \textbf{5.82} & \textbf{56.0} & \textbf{45.3} & \textbf{40.2} \\
\midrule
\multirow{4}{*}{30\%}
 & NaVILA \cite{cheng2024navila}       & 9.00 & 15.1 & 7.50 & 5.20 \\
 & \;\;$+$ FINE                        & \textbf{8.52} & \textbf{21.0} & \textbf{11.9} & \textbf{10.3} \\
 \cmidrule(lr){2-6}
 & InternVLA-N1 \cite{wei2026ground}   & 7.79 & 13.2 & 9.20 & 8.30 \\
 & \;\;$+$ FINE                        & \textbf{7.46} & \textbf{21.3} & \textbf{14.4} & \textbf{13.5} \\
\bottomrule
\end{tabular}
\vspace{-0.2in}
\end{table}
%%%%%%%%%%%%%%%%%%%%%%%%%%%%%%%%%%%%%%%%%%%%%%%%%%%%%%%%%%%%%%%%%%%%%%

%%%%%%%%%%%%%%%%%%%%%%%%%%%%%%%%%%%%%%%%%%%%%%%%%%%%%%%%%%%%%%%%%%%%%%
\begin{table*}[t]
\centering
% \caption{Comparison with published methods on VLN-CE R2R and RxR val-unseen, at
% 100\% training data. Numbers for prior methods are quoted from their
% original publications; observation space and external training data therefore
% differ across rows. Ours uses no external data, JanusVLN${\dagger}$ row uses roughly 10.7M external image-action pairs. Pano.: panoramic RGB; Odo.: odometry; S.RGB: single-view
% monocular RGB. Best per column in \textbf{bold}, second best \underline{underlined};
% -- denotes not reported.}
\caption{VLN-CE R2R and RxR val-unseen results at 100\% training data. Prior results are from original publications and may differ in observation space and external data; InternVLA-N1 DualVLN and NaVILA are evaluated from the authors' released checkpoint. Ours uses no external data \tablefootnote{Refers to VLN
demonstration data; see Sec.~\ref{sec:results} and Table~\ref{tab:ablation} for the auxiliary world model's role.}; JanusVLN${\dagger}$ uses $\sim$10.7M external image-action pairs. Pano.: panoramic RGB; Odo.: odometry; S.RGB: single-view RGB. Best in \textbf{bold}, second best \underline{underlined}; --: not reported.}

\label{tab:sota}
\footnotesize\setlength{\tabcolsep}{4pt}\renewcommand{\arraystretch}{1.05}
\vspace{-0.1in}
\begin{tabular}{l|cccc|cccc|cccc}
\toprule
\multirow{2}{*}{Method} &
\multicolumn{4}{c|}{Observation} &
\multicolumn{4}{c|}{R2R-CE Val-Unseen} &
\multicolumn{4}{c}{RxR-CE Val-Unseen} \\
&
Pano. & Odo. & Depth & S.RGB &
NE$\downarrow$ & OS$\uparrow$ & SR$\uparrow$ & SPL$\uparrow$ &
NE$\downarrow$ & SR$\uparrow$ & SPL$\uparrow$ & nDTW$\uparrow$ \\
\midrule

CMA \cite{hong2022bridging}               & \cmark & \cmark & \cmark &      & 6.20 & 52.0 & 41.0 & 36.0 & 8.76 & 26.5 & 22.1 & 47.0 \\
ETPNav \cite{an2024etpnav}              & \cmark & \cmark & \cmark &      & 4.71 & 65.0 & 57.0 & 49.0 & 5.64 & 54.7 & 44.8 & 61.9 \\
ScaleVLN \cite{wang2023scaling}         & \cmark & \cmark & \cmark &      & 4.80 & -- & 55.0 & 51.0 & -- & -- & -- & -- \\

\midrule

NaVid \cite{zhang2024navid}                  &      &      &      & \cmark & 5.47 & 49.1 & 37.4 & 35.9 & -- & -- & -- & -- \\
UniNaVid \cite{uninavid}              &      &      &      & \cmark & 5.58 & 53.3 & 47.0 & 42.7 & 6.24 & 48.7 & 40.9 & -- \\
FutureNav \cite{zhang2026futurenav}              &      &      &      & \cmark & 5.15 & 61.6 & 55.5 & 51.4 & 5.93 & 53.3 & 45.4 & 59.8 \\
JanusVLN \cite{zeng2026janusvln}              &      &      &      & \cmark & 5.17 & 58.0 & 52.8 & 49.2 & 6.46 & 51.4 & 44.3 & 59.1 \\
JanusVLN${\dagger}$ \cite{zeng2026janusvln}              &      &      &      & \cmark & 4.78 & 65.2 & 60.5 & \textbf{56.8} & 6.06 & 56.2 & 47.5 & 62.1 \\
NaVILA \cite{cheng2024navila}                 &      &      &      & \cmark & 5.22 & 62.5 & 54.0 & 49.0 & 6.77 & 49.3 & 44.0 & 58.8 \\
StreamVLN \cite{wei2025streamvln}                &      &      &      & \cmark & 5.09 & 62.8 & 55.6 & 49.6 & 5.69 & 54.9 & 46.1 & 64.1 \\
InternVLA-N1 DualVLN\cite{wei2026ground}                        &      &      &      & \cmark &
4.83 & 63.3 & 58.2 & 54.0 & 5.91 & 53.5 & 46.1 & 65.3 \\
\midrule
\textbf{InternVLA-N1 DualVLN $+$ FINE (ours)}                        &      &      &      & \cmark &
\textbf{4.29} &
\textbf{65.7} &
\textbf{60.8} &
\underline{55.2} &
\textbf{4.86} &
\textbf{58.0} &
\textbf{47.6} &
\textbf{66.4} \\
\bottomrule
\end{tabular}
\vspace{-0.15in}
% TODO v1 -> DONE
% \revqbox{four items for this table.
% (a)~Check every quoted row against its source: the InternVLA-N1 row
% (4.83/63.3/58.2/52.2) and the NaVILA row (5.20/61.6) differ from the published
% figures. State in the caption which rows we retrained and which are quoted.
% (b)~FutureNav~\cite{zhang2026futurenav} is discussed in Sec.~\ref{sec:rw} but is
% absent here, and it takes the same position as FINE. Add it with checked numbers.
% (c)~The JanusVLN${\dagger}$ row uses roughly 10.7M external image-action pairs. Add an
% external-data column, or add its no-external-data variant, which is the
% comparison this paper argues for.
% (d)~Several 2026 egocentric-RGB methods report higher R2R-CE numbers than any row
% here, so keep this table as context for the backbone and make no claim of state
% of the art.
% Add: fix, janus 0k, futurenav 0k}
\end{table*}
%%%%%%%%%%%%%%%%%%%%%%%%%%%%%%%%%%%%%%%%%%%%%%%%%%%%%%%%%%%%%%%%%%%%%%

\subsection{Experimental Setup}
\label{sec:setup}

\noindent
\textbf{Baselines and evaluation protocol.}
We integrate FINE into InternVLA-N1 DualVLN\cite{wei2026ground} and NaVILA \cite{cheng2024navila} and compare against a diverse set of strong VLN baselines spanning different architectural paradigms.
Table~\ref{tab:sota} reports numbers from the original publications, which use different training data and sensor setups and are thus not directly comparable. Table~\ref{tab:dataeff} and all ablations use our own unified
protocol: same splits, same optimizer, same number of gradient steps, and the same evaluation for every baseline and FINE variant.

\vspace{0.05in}
\noindent
\textbf{Simulation Benchmark Setup.}
We evaluate on validation-unseen set on R2R-CE~\cite{anderson2018vln} and RxR-CE~\cite{rxr}, two continuous VLN-CE benchmarks~\cite{vlnce} built on Matterport3D~\cite{chang2017matterport3d} in Habitat~\cite{savva2019habitat}. R2R-CE contains $\approx$5.6K English trajectories averaging 10\,m, while RxR-CE contains 126K multilingual instructions over longer, 15\,m trajectories.
We report results on the validation-unseen splits to evaluate generalization to unseen environments using standard metrics: Navigation Error (NE, m), Oracle Success Rate (OS), Success Rate (SR), Success weighted by normalized inverse Path Length (SPL), and normalized Dynamic Time Warping (nDTW). 

\noindent
\textbf{Demonstration-budget protocol.}
The 100\%/70\%/30\% 
budgets uniformly subsample training episodes into nested splits
$\mathcal{D}{30}\subset\mathcal{D}{70}\subset\mathcal{D}{100}$.
For R2R, the split sizes are 10,819/7,573/3,245 episodes, while for RxR they are
17,752/12,426/5,326, respectively, then following each model's data processing. The evaluation split and scene coverage are fixed across budgets.
% with
% $|\mathcal{D}{100}|=10{,}819$, $|\mathcal{D}{70}|=7{,}573$, and
% $|\mathcal{D}{30}|=3{,}245$, keeping evaluation split and scene coverage fixed.
% The 100\%/70\%/30\% budgets subsample training \emph{episodes} uniformly at
% random with a seed of 42, nested so that
% $\mathcal{D}_{30}\subset\mathcal{D}_{70}\subset\mathcal{D}_{100}$, holding the
% evaluation split fixed and keeping every training scene represented at each
% budget. 
Backbone pretraining is unchanged across budgets. 
Only the VLN fine-tuning set is subsampled, so we vary the number of task demonstrations, not the scale of the vision-language model. The baseline and FINE use identical subsets at every budget.
% TODO v1 review -> DONE
% \revq{give the absolute episode count per budget and the seed.}

\vspace{0.05in}
\noindent
\textbf{Real-World Evaluation Setup.}
We evaluate on our 30-DoF humanoid robot (1.72 m, 75 kg) equipped with a ZED X Mini camera tilted $30^\circ$ downward. Models run on an RTX 5090 using $\sim$ 20 GB GPU memory, with 95.72ms end-to-end latency and 5 ms overhead over the base model. Monocular RGB is streamed asynchronously, while body-frame commands are transformed using odometry and executed by an MPC controller. We evaluate NaVILA~\cite{cheng2024navila} and InternVLA-N1~\cite{wei2026ground} on three experiments: single-room navigation to visible or occluded landmarks, and room-to-room navigation across multiple visible and occluded landmarks. Landmark and initial positions vary across trials to prevent fixed-path memorization. We run 20 trials per model and experiment and report Success Rate (SR) and Navigation Error (NE).

% Across trials, we vary both the landmark positions and the agent's initial position to verify that the policy navigates toward the instructed landmarks rather than memorizing a fixed path. We run 20 trials per model and experiment, reporting Success Rate (SR) and Navigation Error (NE).

\subsection{Implementation Details}
\label{sec:impl}
\oldtext{For same-scene counterfactuals, the video world model is rolled out for
$H=10$ steps and the final two actions are perturbed. We retain only generated
frames after the counterfactual trajectory diverges from the demonstrated one,
excluding near-duplicate pre-divergence frames from the negative set. 
For the implicit token, we use a two-stage InfoNCE schedule: stage 1 (one epoch) trains with in-batch negatives only, for coarse scene-level discrimination; stage 2 (one epoch) adds hard negatives from the frozen NanoWM~\cite{nanowm} world model, rolled out with the trajectory's last 2
actions heading-flipped to produce same-scene counterfactual futures that differ only near the episode endpoint. NanoWM is fine-tuned on the same R2R-CE train split used for FINE, so hard negatives
stay within the supervised distribution with no val/test leakage.

For explicit supervision, frozen VGGT~\cite{vggt} provides 2048-D patch features aligned with \(f_{\mathrm{enc}}\) at patch size 14, requiring no resampling. 
Within each budget, baseline and FINE use the same subset and epochs, yielding equal optimizer updates. Reduced-data budgets therefore receive fewer updates than 100\%.
Following Eq.~\eqref{eq:total}, we use a two-phase schedule: \((w_a,w_e,w_i)=(0.3,1.0,1.0)\) for the first half of training and \((1.0,0.3,0.3)\) for the second half. We train on 8 H100 GPUs with peak learning rates of \(2\times10^{-4}\) for the language model and \(5\times10^{-6}\) for the vision encoder. Supervision targets are produced by a separate frozen vision encoder, with no gradient flow through the target branch.}

\newtext{
For implicit supervision, we use a two-stage InfoNCE schedule. Stage~1
(one epoch) uses in-batch negatives for coarse discrimination, while Stage~2
(one epoch) adds same-scene counterfactuals generated by a frozen
video world model NanoWM~\cite{nanowm}. Specifically, NanoWM rolls out $H=10$ steps from the
current observation with the final two actions heading-flipped; only
post-divergence frames are retained as hard negatives, excluding near-duplicate
frames before the trajectories diverge. NanoWM is fine-tuned only on the
corresponding R2R-CE training split, avoiding validation/test leakage.

For explicit supervision, frozen VGGT~\cite{vggt} provides 2048-D geometric
patch features aligned with the semantic encoder at patch size 14, requiring no
resampling. Target features are produced by frozen encoders with no gradient
flow through the supervision branches. Following Eq.~\eqref{eq:total}, we use
$(w_a,w_e,w_i)=(0.3,1.0,1.0)$ in the first half of training and
$(1.0,0.3,0.3)$ in the second half. Baseline and FINE use identical data
subsets and training epochs within each demonstration budget; consequently,
reduced-data settings receive fewer optimizer updates than the full-data
setting. We train on 8 H100 GPUs with peak learning rates of
$2\times10^{-4}$ for the language model and $5\times10^{-6}$ for the vision
encoder.}

\subsection{Main Results}
\label{sec:results}
\oldtext{We evaluate FINE as a training-time adaptation framework under three
demonstration budgets, 100\%, 70\% and 30\%, on two backbones representing
different policy classes: NaVILA for discrete actions and InternVLA-N1 for
continuous trajectory generation. Each baseline is retrained under the same
budget and setup before FINE is integrated.
As shown in Table~\ref{tab:dataeff}, FINE improves the underlying policy at every
budget measured, and by a larger margin as demonstrations are withheld. On
InternVLA-N1, FINE adds 2.6 SR points (58.2 to 60.8) and 1.2 SPL points (54.0 to
55.2) at full data; at the 70\% budget the margins widen to 6.8 and 5.8 points
($+17.7\%$ and $+16.9\%$ relative). Read as a recovery fraction, reducing
demonstrations from 100\% to 70\% costs the baseline 19.7 SR points, of which
FINE returns 6.8, about a third. 
NaVILA exhibits the same pattern where measured:
FINE improves SR/SPL by $1.6/1.7$ points at 100\% and $5.5/6.0$ points
at 70\%.}

\newtext{\noindent
\textbf{Data-efficient adaptation.} We evaluate FINE under 100\%, 70\%, and 30\% demonstration budgets on NaVILA
(discrete actions) and InternVLA-N1 (continuous trajectories), using matched
training subsets and protocols within each budget.

As shown in Table~\ref{tab:dataeff}, FINE improves both backbones at all budgets,
with larger gains as demonstrations become scarce. On InternVLA-N1, FINE
improves SR/SPL by $2.6/1.2$ points at full data and $6.8/5.8$ points at the
70\% budget. The latter recovers roughly one third of the 19.7-point SR drop
caused by reducing the training set. NaVILA shows the same trend, with SR/SPL
gains increasing from $1.6/1.7$ to $5.5/6.0$ points.}

These results indicate that future-informed supervision is
particularly beneficial when demonstrations are limited, while the remaining
gap to full-data performance shows that FINE complements rather than replaces
additional training data.

\vspace{0.05in}
\noindent
\oldtext{\textbf{Benchmark results on VLN-CE.}Table~\ref{tab:sota} places FINE among published VLN-CE methods at full
data. Using monocular RGB alone, FINE improves its InternVLA-N1 backbone on every
metric of both benchmarks: on R2R-CE from 58.2 to 60.8 SR and 54.0 to 55.2 SPL,
with NE reduced from 4.83\,m to 4.29\,m; on RxR-CE, which is longer and
multilingual, by 4.5 SR points (53.5 to 58.0), 1.5 SPL points and 1.1 nDTW. No external demonstrations are used; the auxiliary world model contributes a refinement on top of an already-positive coarse-negative result, isolated in Table~\ref{tab:ablation}. It remains competitive with methods that consume panoramic views, depth and
odometry. JanusVLN retains the best R2R-CE SPL, 56.8 against 55.2, with a large volume of additional image-action pairs; on RxR-CE FINE
leads it on all reported metrics. SR improves more than SPL on RxR-CE, so the
additional successes come at some cost in path length, while the nDTW gain
indicates that the extra length stays route-consistent. We do not claim a
leaderboard position: concurrent egocentric-RGB methods report stronger absolute
numbers with larger backbones and their own training recipes. Our claim is the
gain a fixed backbone obtains per demonstration, which Table~\ref{tab:dataeff}
isolates.}

\newtext{\noindent\textbf{Benchmark results on VLN-CE.}
Table~\ref{tab:sota} compares FINE with published VLN-CE methods at full
training data. Using monocular RGB only, FINE improves InternVLA-N1 on every
reported metric: on R2R-CE, SR increases from 58.2 to 60.8 and NE decreases
from 4.83\,m to 4.29\,m; on the longer, multilingual RxR-CE benchmark, FINE
gains 4.5 SR points (53.5 to 58.0), together with improvements in SPL and nDTW.
These gains are achieved without external demonstrations and remain competitive
with methods using richer inputs such as panoramic RGB, depth, or odometry.
Although published results are not directly comparable due to different
backbones and training recipes, the consistent improvement over a fixed
InternVLA-N1 backbone supports the effectiveness of FINE's future-informed
supervision.}

%%%%%%%%%%%%%%%%%%%%%%%%%%%%%%%%%%%%%%%%%%%%%%%%%%%%%%%%%%%%%%%%%%%%%%
\begin{table}[t]
\centering
\caption{Real-world evaluation with SR (\%) $\uparrow$ and NE (m) $\downarrow$. Experiments 1–2: single-room (visible and occluded landmarks); Experiment 3: room-to-room (multiple landmarks).}
\label{tab:realworld}
\footnotesize\setlength{\tabcolsep}{4pt}\renewcommand{\arraystretch}{1.05}
\begin{tabular}{llccc}
\toprule
\textbf{Model} & \textbf{Data} & \textbf{Experiment 1} & \textbf{Experiment 2} & \textbf{Experiment 3} \\
\midrule
NaVILA                  & 100\% & 60 / 2.3  & 10 / 8.8 & 0 / 10 \\
\;\;$+$ FINE            & 100\% & 65 / 2.2 & 15 / 7.2 & 10 / 8.5 \\
InternVLA-N1            & 100\% & 80 / 0.4 & 30 / 3.5 & 30 / 5.5 \\
\;\;$+$ FINE            & 100\% & \textbf{85} / \textbf{0.4} & \textbf{60} / \textbf{2.3} & \textbf{50} / \textbf{4.2} \\
InternVLA-N1            & 70\%  & 60 / 1.7 & 10 / 6.4 & 10 / 6.7 \\
\;\;$+$ FINE            & 70\%  & 75 /  1.5& 25 / 3.7 & 25 / 4.2 \\
\bottomrule

\end{tabular}
\end{table}
%%%%%%%%%%%%%%%%%%%%%%%%%%%%%%%%%%%%%%%%%%%%%%%%%%%%%%%%%%%%%%%%%%%%%%

\begin{figure}[t]
    \centering
    \includegraphics[width=0.98\linewidth]{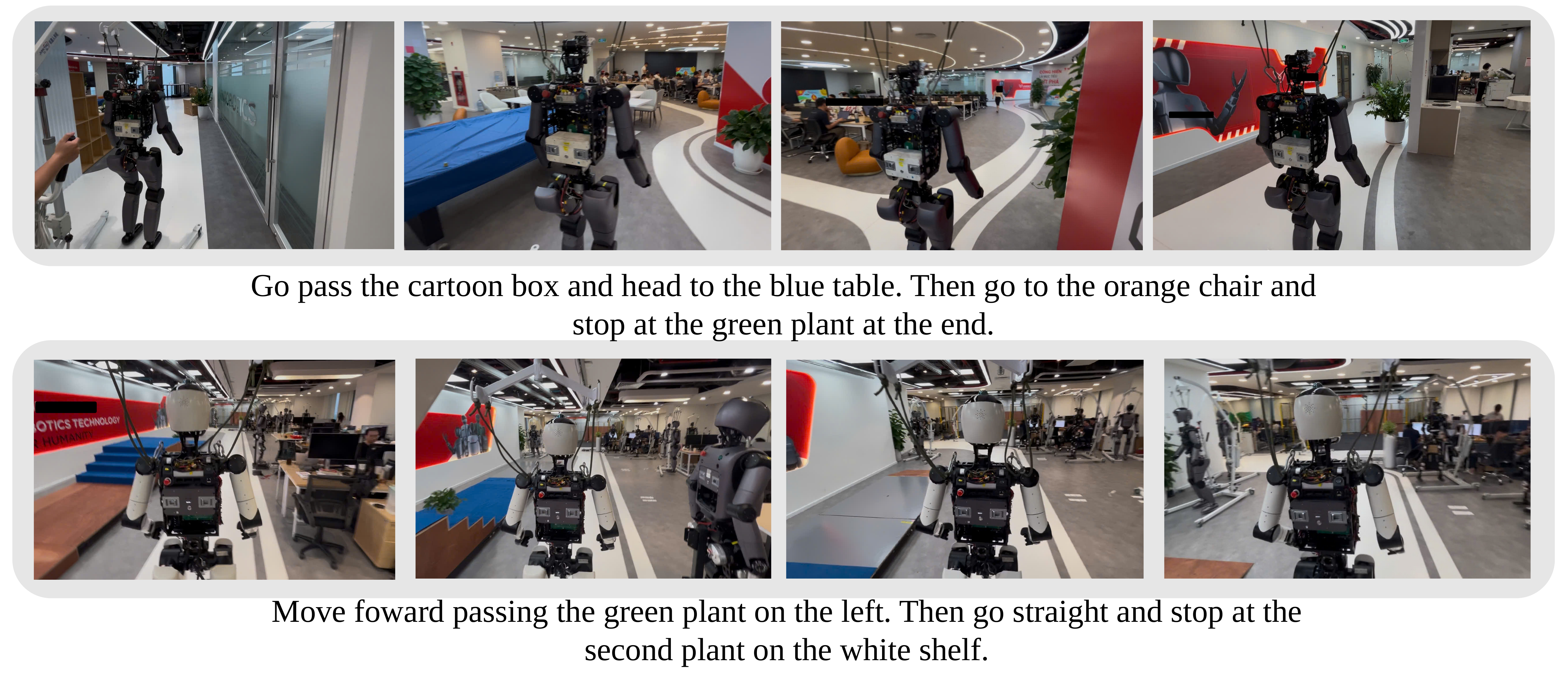}
    \caption{Qualitative results from the real-world deployment of FINE. Given an  instruction, the robot moves through different areas of the workspace and stops at the specified goal.}
    \vspace{-2 mm}
    \label{fig:qualitative}
    \vspace{-3 mm}
\end{figure}

\oldtext{\noindent
\textbf{Real-world navigation.}
Table~\ref{tab:realworld} summarizes the real-world performance across the three experiments. Across trials, we randomly vary landmark locations and the agent's initial position while providing prompts that specify only the target landmarks, as illustrated in Fig.~\ref{fig:realworld} and Fig.~\ref{fig:qualitative}, without additional directional guidance such as turning left or right. Under this setup, FINE achieves the strongest overall results with InternVLA-N1, with more pronounced gains under landmark occlusion and longer-horizon navigation. Specifically, at a 100\% data budget, FINE improves SR by 30 percentage points in \textit{Experiment 2} and 20 percentage points in \textit{Experiment 3}, while achieving a 15 percentage-point improvement at a 70\% data budget. Improvements on NaVILA are smaller, partly due to its weaker baseline performance. In contrast, the gains are limited in \textit{Experiment 1}, where the target landmark is directly visible within a single room.
Qualitative results on our internal robots are shown in Fig.~\ref{fig:realworld} and Fig.~\ref{fig:qualitative} . In particular, Fig.~\ref{fig:realworld} shows from the top-down localization map that the robot changes its orientation after Step 5, but is still able to re-anchor to the final landmark and continue along the intended trajectory.}

\vspace{0.05in}
\newtext{\noindent\textbf{Real-world navigation.}
Table~\ref{tab:realworld} reports real-world results across three navigation
settings with randomized landmark and initial robot positions. Prompts specify
only the target landmarks, without directional cues such as \emph{turn left}
or \emph{turn right}. FINE yields its largest gains when landmarks are occluded
or navigation spans multiple landmarks. With InternVLA-N1 at 100\% data, FINE
improves SR by 30 points in \textit{Experiment 2} and 20 points in
\textit{Experiment 3}; at 70\% data, it improves SR by 15 points in both
settings. Gains are smaller in \textit{Experiment 1}, where the target is
directly visible, and on the weaker NaVILA backbone.
Fig.~\ref{fig:realworld} and~\ref{fig:qualitative} further illustrate that
FINE can re-anchor to instruction-relevant landmarks after intermediate
viewpoint changes and continue along the intended route.}

%%%%%%%%%%%%%%%%%%%%%%%%%%%%%%%%%%%%%%%%%%%%%%%%%%%%%%%%%%%%%%%%%%%%%%
\begin{table}[t]
\centering
\caption{Contribution of the two FINE branches on R2R-CE val-unseen, at 100\% and 70\% demonstrations.}
\label{tab:ablation_FINE}
\footnotesize\setlength{\tabcolsep}{4pt}\renewcommand{\arraystretch}{1.05}
\begin{tabular}{lcccccccc}
\toprule
& \multicolumn{4}{c}{100\%} & \multicolumn{4}{c}{70\%} \\
\cmidrule(lr){2-5} \cmidrule(lr){6-9}
\textbf{Method}
& \textbf{NE$\downarrow$} & \textbf{OS$\uparrow$} & \textbf{SR$\uparrow$} & \textbf{SPL$\uparrow$}
& \textbf{NE$\downarrow$} & \textbf{OS$\uparrow$} & \textbf{SR$\uparrow$} & \textbf{SPL$\uparrow$} \\
\midrule
Baseline
& 4.83 & 63.3 & 58.2 & 54.0
& 6.45 & 46.0 & 38.5 & 34.4 \\
$+$ explicit only
& 4.38 & 64.4 & 59.5 & 53.3
& 5.96 & 53.8 & 44.4 & 38.3 \\
$+$ implicit only
& 4.46 & 65.7 & 59.6 & 54.4
& 5.58 & 48.6 & 42.5 & 37.4 \\
\midrule
$+$ FINE (full)
& \textbf{4.29} & \textbf{65.7} & \textbf{60.8} & \textbf{55.2}
& \textbf{5.58} & \textbf{56.0} & \textbf{45.3} & \textbf{40.2} \\
\bottomrule
\end{tabular}
\end{table}

\begin{table}[t]
\centering
\caption{Design choices on R2R-CE val-unseen at the 70\% budget (Table~IV
baseline: 38.5 SR). \textsc{Base} marks rows built on the \emph{Expl.}
(explicit-only) or \emph{Impl.}/\emph{Full} configuration; see text for details.}
\label{tab:ablation}
\footnotesize\setlength{\tabcolsep}{4pt}\renewcommand{\arraystretch}{1.0}
\begin{tabular}{llcccc}
\toprule
\textbf{Base} & \textbf{Variant} & \textbf{NE}$\downarrow$ & \textbf{OS}$\uparrow$ & \textbf{SR}$\uparrow$ & \textbf{SPL}$\uparrow$ \\
\midrule
       & $S = 2$ slots                     & 6.21 & 47.0 & 40.0 & 36.0 \\
Expl.  & $S = 10$ slots                    & \textbf{5.96} & \textbf{53.8} & \textbf{44.4} & \textbf{38.3} \\
       & $S = 18$ slots                    & 5.96 & 49.0 & 41.3 & 36.3 \\
\midrule
\multirow{3}{*}{Expl.} & Semantic only            & 6.04 & 48.8 & 42.0 & 37.2 \\
                        & Geometric only           & 5.96 & 50.0 & 42.9 & 37.7 \\
                        & Semantic $+$ Geometric   & \textbf{5.96} & \textbf{53.8} & \textbf{44.4} & \textbf{38.3} \\
\midrule
\multirow{2}{*}{Impl.} & Cross-environment negatives only & 5.58 & 47.5 & 40.5 & 36.3 \\
                        & $+$ World-model rollouts         & \textbf{5.58} & \textbf{48.6} & \textbf{42.5} & \textbf{37.4} \\
\midrule
Impl.                  & Shuffled-future placebo          & 5.95 & 40.08 & 34.91 & 30.8 \\
\multirow{2}{*}{Full}  & W/o token-isolation mask         & \textbf{5.71} & 51.1 & 41.8 & \textbf{38.4} \\
                        & Random explicit initialization   & 5.89 & \textbf{51.6} & \textbf{42.3} & 37.8 \\
\bottomrule
\end{tabular}
\end{table}
%%%%%%%%%%%%%%%%%%%%%%%%%%%%%%%%%%%%%%%%%%%%%%%%%%%%%%%%%%%%%%%%%%%%%%
\subsection{Ablation Studies}

\noindent
\textbf{Contribution of the two branches.}
Table~\ref{tab:ablation_FINE} decomposes FINE at two budgets. Combining both
branches improves NE, SR and SPL over either branch alone and matches the
stronger branch on OS. The combined gain is sub-additive: at the 70\% budget
the explicit branch alone contributes 5.9 SR points and the implicit branch 3.0,
while together they contribute 6.8. The two supervision signals therefore overlap
in part, which follows from their construction, since both are computed on the
same future landmark region and differ in how they use it, not in what they
observe. The explicit branch carries most of the gain, and the implicit branch
adds to it in every metric.

\vspace{0.05in}
\noindent
\oldtext{\textbf{Design choices.}
Table~\ref{tab:ablation} reports the individual design decisions. A moderate
number of landmark slots works best: $S{=}10$ outperforms both $S{=}2$, which
cannot represent all instruction landmarks, and $S{=}18$, whose surplus slots are
empty on most instructions. The non-monotonicity is informative in itself, since
a gain driven purely by added capacity would not fall again at $S{=}18$. The
remaining rows compare the two explicit targets against each other, the two
negative sets, the landmark-region against the full-scene implicit target, and
three controls: a placebo in which the future targets are drawn from a different
episode, removal of the token-isolation mask, and random initialization of the
explicit slots instead of object-aware initialization.}

\newtext{\noindent\textbf{Design choices.}
Table~\ref{tab:ablation} examines which components make future supervision
effective. A moderate number of landmark slots performs best:
$S{=}10$ outperforms both $S{=}2$, which under-represents multi-landmark
instructions, and $S{=}18$, where many slots are unused. This non-monotonic trend
suggests that the gain comes from matching the structure of the instruction,
rather than simply increasing model capacity.
The remaining ablations support three further conclusions. First, semantic and
geometric targets are complementary, indicating that recognizing \emph{what}
landmark appears and \emph{how} it is spatially arranged provide distinct
signals. Second, same-scene counterfactuals are more informative than
cross-episode negatives alone because they require route-level, rather than
scene-level, discrimination. Finally, performance drops when future targets are
shuffled, token isolation is removed, or landmark slots are randomly
initialized, showing that the gains depend on correctly structured future
supervision rather than an additional auxiliary loss alone.}

\vspace{0.05in}
\noindent
\oldtext{\textbf{Role of world-model rollouts.}
The shuffled-future placebo fails to converge, so the objective acts as noise that degrades model performance. With cross-environment negatives, model performance recovers, and NanoWM rollouts add 2.0 further SR points at unchanged NE. This progression shows that what matters is whether the target is correct -NanoWM's gain comes from sharpening a signal that is already working.}

\newtext{\noindent\textbf{Role of world-model rollouts.}
The shuffled-future placebo fails to converge, showing that incorrect future
targets act as noise rather than useful supervision. Cross-episode negatives
recover performance, and NanoWM same-scene counterfactuals add a further
$+2.0$ SR at unchanged NE. This indicates that gains depend on both target
correctness and negative difficulty, with world-model rollouts sharpening
route-level discrimination among plausible futures.}

%%%%%%%%%%%%%%%%%%%%%%%%%%%%%%%%%%%%%%%%%%%%%%%%%%%%%%%%%%%%%%%%%%%%%%
%%%%%%%%%%%%%%%%%%%%%%%%%%%%%%%%%%%%%%%%%%%%%%%%%%%%%%%%%%%%%%%%%%%%%%

\section{Conclusion}
\oldtext{We presented FINE, a future-informed navigation encoding that recovers trajectory supervision overlooked by standard observation-to-action learning. FINE attaches two
auxiliary token groups to an unmodified VLN backbone: explicit landmark slots
that regress the semantic and geometric features of the upcoming landmark region,
and an implicit token that separates the realized future landmark state from
same-scene world-model counterfactuals. Both are supervised entirely by frames
already contained in each demonstration, and every future-dependent module is
removed after training, so the deployed policy performs no rollout and observes
no future.
On R2R-CE and RxR-CE val-unseen, FINE improves its backbone on every metric of
both benchmarks, and the margin widens as demonstrations are withheld: at a 70\%
budget it recovers about a third of the success rate lost to the reduction. The
policy 
transfers effectively to humanoid robot navigation across three difficulty levels.}

\newtext{This work suggests a simple but useful principle for data-efficient VLN:
\emph{a demonstration should supervise more than the action it contains}.
Future observations already reveal which instruction-relevant landmarks are
eventually encountered, what they look like, and how they are spatially
structured. FINE shows that extracting this information during training can
improve navigation without requiring future prediction or rollout for action
selection at deployment.

Our results highlight three main takeaways. First, future supervision is most
effective when it is \emph{selective}: focusing on instruction-relevant landmark
regions is more useful than treating the future scene as an undifferentiated
target. Second, semantic appearance, 3D geometry, and route-sensitive
counterfactuals provide complementary signals for learning navigation
representations. Third, the benefit grows as demonstrations become scarce and
when landmarks are occluded or navigation horizons become longer. At a 70\%
R2R-CE demonstration budget, FINE recovers roughly one third of the success-rate
drop caused by reducing the training data, while also transferring to real-world
humanoid navigation.
More broadly, these findings motivate viewing future observations as
\emph{training-time teachers}: information already present in embodied
trajectories can be converted into richer supervision, improving how much is
learned from each demonstration without increasing deployment-time planning
complexity.}

\vspace{0.05in}
\oldtext{\noindent\textbf{Limitations.}
FINE assumes instructions can be parsed into ordered, detectable open vocabulary landmarks and that the target landmark appears in the supervising future frame; otherwise, the auxiliary loss is skipped.
Implicit negative quality is bounded by the fidelity of the world model, which was not trained on indoor egocentric navigation video.
Training uses three frozen auxiliary models (a detector, geometric teacher, and world model) that are removed at inference but add training-time overhead.
Our budgets subsample demonstrations from fixed training scenes, so results
reflect demonstration efficiency rather than environment diversity.
Training is epoch-matched across budgets; consequently, reduced-data runs
receive fewer optimization steps than the full-data setting.}

\newtext{\noindent \textbf{Limitations and future directions.}
FINE relies on instructions that can be decomposed into ordered, detectable
landmarks and on those landmarks appearing in future observations. Its training
also depends on auxiliary models, including a detector, geometric teacher, and
video world model, so supervision quality is partly bounded by these components.
Finally, our reduced-data study measures demonstration efficiency within fixed
training environments, rather than generalization across broader scene
diversity. Extending FINE to more abstract navigation cues, lighter supervision,
and more diverse environments is an important direction.}

% \section*{Acknowledgment}
% The authors used ChatGPT to assist with language polishing and figure improvement, and Claude to support the development of training and evaluation code.
% All AI-assisted outputs were critically reviewed, verified, and edited by the authors, who take full responsibility for the accuracy, integrity, and originality of the final content.

\bibliographystyle{IEEEtran}
\bibliography{references}

\end{document}